\documentclass[letterpaper, 10 pt, conference]{ieeeconf}  

\usepackage{amsmath,amsfonts,amssymb}
\usepackage{graphicx}
\usepackage[colorlinks=true, allcolors=blue]{hyperref}
\usepackage{multirow}
\usepackage{booktabs}
\usepackage{subfigure}
\usepackage{bbding}
\usepackage{threeparttable}

\usepackage{caption}

\usepackage{cite} 

\usepackage{amssymb}
\usepackage{amsbsy}

\usepackage{booktabs} 
\usepackage{makecell}
\usepackage{multirow}
\usepackage{overpic}
\usepackage{diagbox}

\usepackage{adjustbox}
\usepackage{array}

\usepackage{verbatim}

\usepackage{color, soul}
\sethlcolor{blue}

\newcommand{\hlr}[1]{{\color{red}{#1}}}
\newcommand{\hlb}[1]{{\color{blue}{#1}}}

\IEEEoverridecommandlockouts                              

\title{\LARGE \bf
Traffic Context Aware Data Augmentation for Rare Object Detection in Autonomous Driving
}

\author{Naifan Li, Fan Song$^{\dagger}$, Ying Zhang, Pengpeng Liang and Erkang Cheng$^{*}$ 
	\thanks{Naifan Li, Fan Song, Ying Zhang, and Erkang Cheng are with NullMax, Shanghai, 201210, China. Pengpeng Liang is with School of Information Engineering, Zhengzhou University, 450001, China.
		{\tt\small linaifan, songfan, zhangying, chengerkang@nullmax.ai; liangpcs@gmail.com}}
	\thanks{$\dagger$ Work done during an internship at NullMax.}
	\thanks{*Corresponding author.}
}

\begin{document}

\maketitle
\thispagestyle{empty}
\pagestyle{empty}

\begin{abstract}
Detection of rare objects (e.g., traffic cones, traffic barrels and traffic warning triangles) is an important perception task to improve the safety of autonomous driving. Training of such models typically requires a large number of annotated data which is expensive and time consuming to obtain. To address the above problem, an emerging approach is to apply data augmentation to automatically generate cost-free training samples. In this work, we propose a systematic study on simple Copy-Paste data augmentation for rare object detection in autonomous driving. Specifically, local adaptive instance-level image transformation is introduced to generate realistic rare object masks from source domain to the target domain. Moreover, traffic scene context is utilized to guide the placement of masks of rare objects. To this end, our data augmentation generates training data with high quality and realistic characteristics by leveraging both local and global consistency. In addition, we build a new dataset, Rare Object Dataset (ROD), consisting 10k training images, 4k validation images and the corresponding labels with a diverse range of scenarios in autonomous driving. 
Experiments on ROD show that our method achieves promising results on rare object detection. We also present a thorough study to illustrate the effectiveness of our local-adaptive and global constraints based Copy-Paste data augmentation for rare object detection.
The data, development kit and more information of ROD are available online at: \url{https://nullmax-vision.github.io}.
\end{abstract}

\section{INTRODUCTION}

\begin{figure}[t]
	\begin{center}
		\begin{tabular}
			{@{\hspace{.0mm}}c@{\hspace{.2mm}} @{\hspace{.2mm}}c@{\hspace{.0mm}}}
			\vspace{-2pt}
			\includegraphics[width=0.45\linewidth]{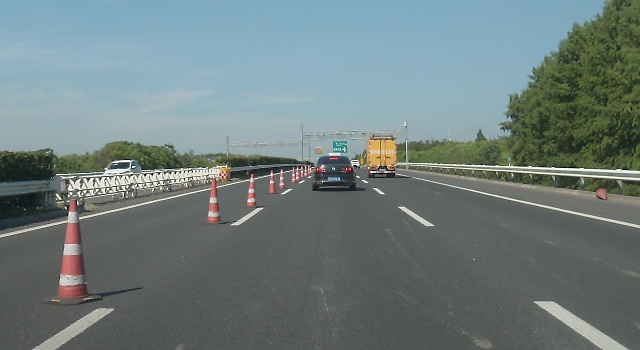}&
			\includegraphics[width=0.45\linewidth]{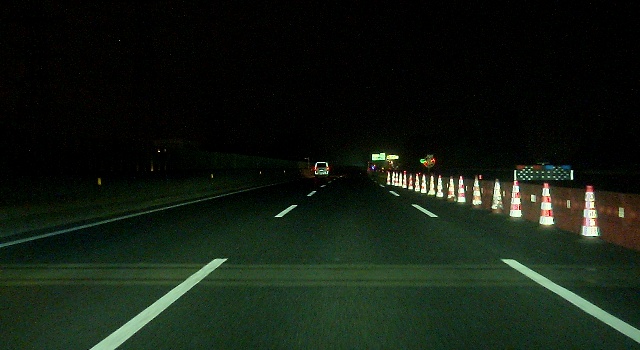} \\
			\vspace{-2pt}
			\includegraphics[width=0.45\linewidth]{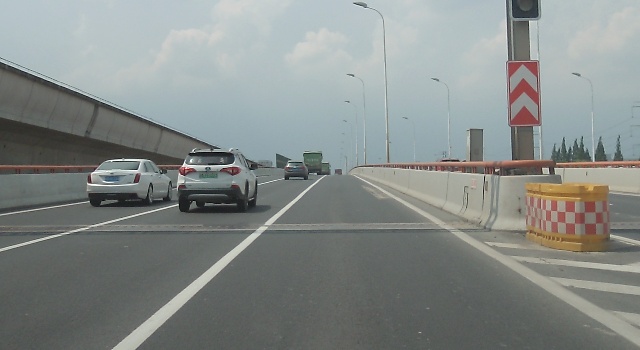}&
			\includegraphics[width=0.45\linewidth]{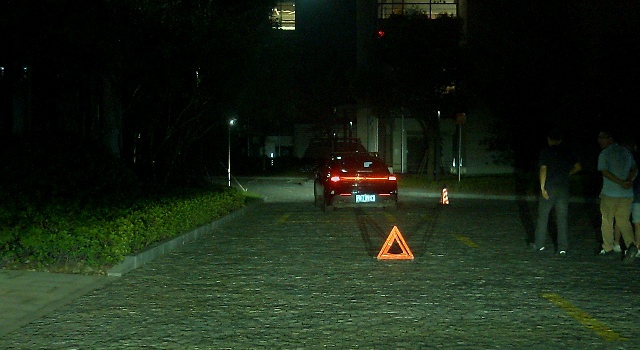} \\
		\end{tabular}
	\end{center}
	{\caption{Sample images of rare objects (traffic cones, traffic barrels and traffic warning triangles) in autonomous driving.} \label{fig:rare_objects}}
\end{figure}

Camera as a cost-effective sensor, is one of the main sensor modalities used for the perception system in autonomous driving. Object detection in images is a critical component of such perception systems to identify objects or obstacles which are the key features of the environment around an ego car. Compared to common objects such as road users including vehicles, pedestrians, etc, rare object detection like traffic cones, traffic barrels and traffic warning triangles, as shown in Fig.~\ref{fig:rare_objects}, is also an important perception task to improve the safety of autonomous driving. For example, detection of these rare objects is essential to prevent traffic accidents or car crashes.

Recently, deep learning based detectors such as Faster-RCNN~\cite{ren2016faster}, SSD~\cite{liu2016ssd}, FCOS~\cite{tian2020fcos}, YOLO~\cite{yolov1} and its variants~\cite{redmon2017yolo9000, redmon2018yolov3, glenn_jocher_2020_4154370}, have achieved remarkable performance. Training such models usually requires a large amount of annotated data, which is expensive and time consuming for both data collection and data annotation. Therefore, a major challenge in rare object detection using these detectors is the lack of large annotated datasets. For example, finding a large labeled dataset similar to common object categories (e.g., vehicle, pedestrians) is unlikely since rare objects are extremely less frequently present in the traffic scene.

For object detection task with less training samples, it can be treated as object detection with an unbalanced or long-tail dataset. Several existing works tackle the problem by re-sampling training data~\cite{cui2019class} and re-weighting loss function~\cite{menon2020long, li2020overcoming}. Although these approaches can boost the performance, they usually need to carefully finetune the hyperparameters of sample rate or loss weight, which heavily rely on expert experience. 

Another line of research to solve the data scarcity challenge is to apply data augmentation. The goal of data augmentation is to generate large annotated training samples with minimal effort. Recent data augmentation approaches for object detection can be roughly divided into image-level and instance-level augmentation. For example, image-level data augmentation approaches include geometric transformation (e.g., random flipping, random cropping)~\cite{lin2017focal}, photometric transformation (e.g., color jittering) \cite{glenn_jocher_2020_4154370}, and image occlusion (e.g., Random Erase~\cite{zhong2020random}, Cutout~\cite{devries2017improved}). The image-level data augmentation is effective when there is a certain amount of image-level training data which contains a specific object category. In the situation there is no such image-level training data, instance-level data augmentation is a way to solve the problem. For example, Copy-Paste~\cite{cut_paste} is a common instance-level data augmentation which copies instance masks of a specific object category from source domain and paste them to the target domain. Some works also model context information to guide the placement of the instance masks in the target images~\cite{dvornik2018modeling, dvornik2019importance}. In contrast to the necessity of modeling the context for the placement of instance masks, study in~\cite{cut_paste} shows that local consistency is more important than where to paste.
However, apart from general object detection, data augmentation for rare object detection in autonomous driving has not been explored.

In this paper, we provide a systematic study on data augmentation for rare object detection in autonomous driving. We introduce a simple and effective Copy-Paste data augmentation technique to solve the challenges of data scarceness for rare object detection in autonomous driving. 
Specifically, we apply instance-level transformations of object masks from source domain, such as color transformation and scaling to create realistic object instances. Furthermore, 
traffic scene context (e.g., freespace, common objects and traffic lanes) obtained from a multi-task deep model are recognized as global geometry constraints. This global context is used to guide the placement of the instance masks. 
By pasting the local-adaptive instance masks to target images under the guidance of global context, one can generate a large amount of training data of rare objects for autonomous driving. To this end, our data augmentation ensures the training data with high quality and realism by leveraging both local and global consistency. 

To evaluate our data augmentation for rare object detection in autonomous driving, we have built and released a new Rare Object Dataset (\textbf{ROD}). The dataset contains 10k training images, 4k validation images. The dataset also covers a diverse range of scenarios including different road grades , weather conditions and various capture time.

In the experiments, we mainly focus on traffic cone detection. Our proposed data augmentation method yields promising results on traffic cone detection. 
We also show that our method is able to generalize to other types of rare object detection tasks. 
Additionally, the effect of components of our method is extensively studied in the ablation experiments. We analyze the sensitivity of our method to domain of instance mask, number of instance masks and number of augmented training images. The experimental results validate that our method is efficient to provide training images for rare object detection.

\section{RELATED WORK}

\begin{figure*}[htb]
    \vspace{2mm}
	\centering
	\includegraphics[width=0.9\linewidth]{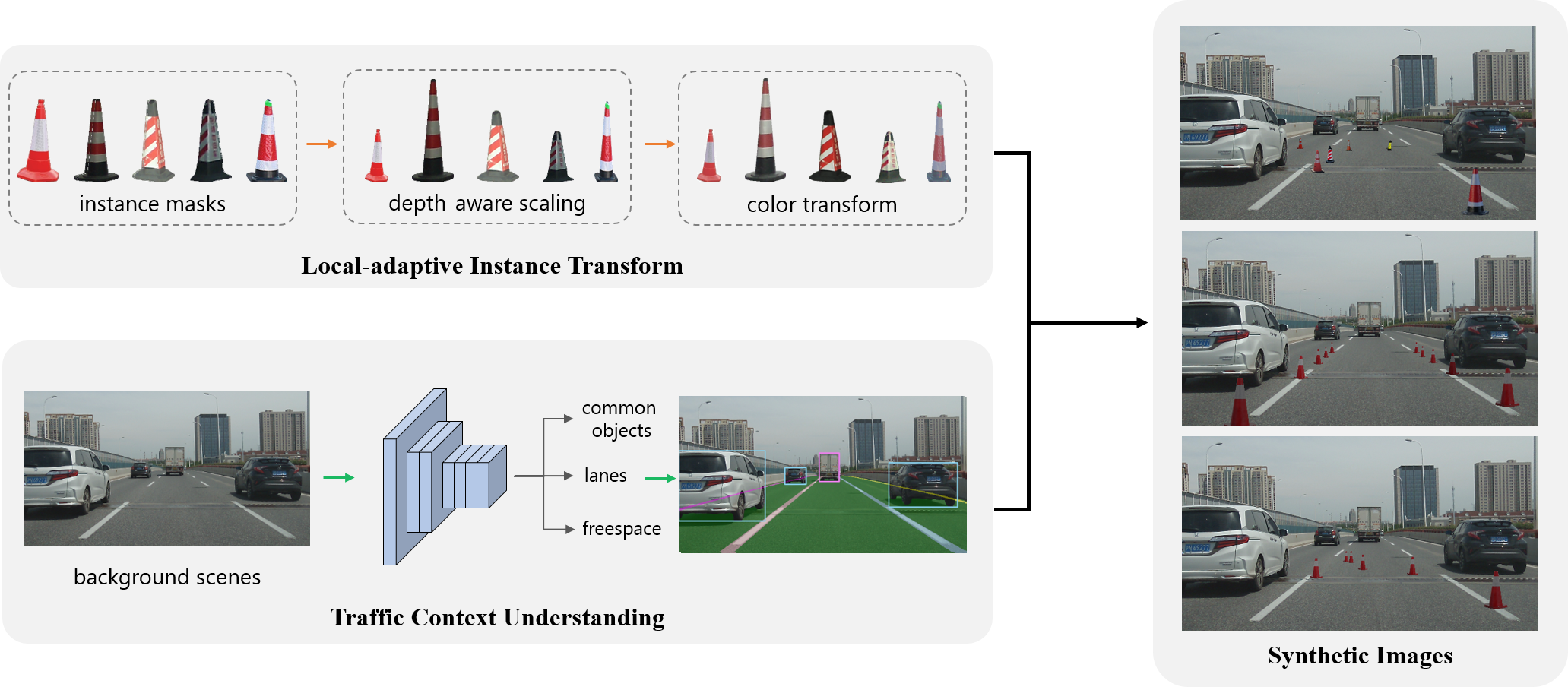}
	\caption{Overview of our data augmentation method. We apply local-adaptive instance transform of rare object (e.g., traffic cone) masks and paste them into background images. Traffic scene context is used as global constraints to guide the placement of instance masks.}
	\label{fig:overview}
\vspace{-6mm}
\end{figure*}

\textbf{Object Detection}  Advanced deep learning detectors can be categorized into anchor-based and anchor-free detectors. The former one can also be grouped to two-stage and one-stage methods.

Faster-RCNN~\cite{ren2016faster}, a dominating two-stage anchor-based detector, consists of a separate region proposal subnetwork (RPN) to generate ROI proposals and a second subnetwork is applied to classify and regress the ROIs to compute the detection results. In contrast, one-stage detectors such as SSD~\cite{liu2016ssd} and YOLO series~\cite{redmon2017yolo9000, redmon2018yolov3, yolov4, glenn_jocher_2020_4154370} simultaneously predict multiple bounding boxes using a single CNN network to achieve high computational efficiency.
In anchor-based detectors, pre-defined anchors are incorporated for proposal generation (two-stage detector) or bounding box regression (one-stage detector). Different to anchor-based approaches, anchor-free detectors directly compute object detection results without pre-defined anchors. Such detectors can be divided into keypoint-based methods such as CornerNet~\cite{law2018cornernet}, ExtremeNet~\cite{zhou2019bottom}, and CenterNet~\cite{zhou2019objects} and center-based methods like FCOS \cite{tian2020fcos}.

In this paper, we apply YOLOv5 \cite{glenn_jocher_2020_4154370}, an one-stage anchor-based detector, to study data augmentation for rare object detection for autonomous driving.

\textbf{Rare Object Detection} Different from common object detection, rare object detection generally has less frequency training samples of specific categories in the dataset. Therefore, it can be treated as object detection with an unbalanced or long-tail dataset.

To tackle the long-tail learning problem, data re-sampling~\cite{cui2019class} is a common solution to balance the data distribution. 
It typically oversamples training data from tail categories while undersamples those from head categories.
Similarly, loss re-weighting~\cite{menon2020long, li2020overcoming} is another strategy to alleviate the problem by adjusting weights for different categories. 
For example, it assigns large weights for tail classes while assigns small weights for head classes.
Both resampling and reweighting approaches rely on adhoc settings to finetune hyperparameters of the networks.
In addition, these approaches still require a certain number of training samples. It is difficult to collect adequate data for rare objects in autonomous driving.

\textbf{Data Augmentation} Recent studies show that data augmentation can lead to significant performance gain in various computer vision tasks. Data augmentation approaches can be grouped into geometric transformation augmentation and photometric augmentation, which are usually for  image-level data augmentation. For example, geometric transformations~\cite{lin2017focal} such as random flipping, random scaling, random cropping and photometric augmentation \cite{glenn_jocher_2020_4154370} (e.g., hue, brightness, contrast, sharpness and saturation) have played a big role in achieving state-of-the-art results on visual tasks. 

To address partial occlusion problem, occlusion-based data augmentation as Random Erasing~\cite{zhong2020random}, Hide-And-Seek~\cite{singh2017hide} and Cutout~\cite{devries2017improved} have been explored. In order to alleviate the information loss caused by these occlusion augmentation methods, CutMix~\cite{yun2019cutmix}, the combination of Cutout~\cite{devries2017improved} and Mixup~\cite{zhang2018mixup}, achieves a noticeable gain in performance of image classification task.

Unlike aforementioned methods which mainly performed on a single image, the Mosaic data augmentation~\cite{yolov4} randomly selects 4 images from the train set and stitches them a synthetic image and can largely boost the performance.

\textbf{Copy-Paste Augmentation} Different to image-level data augmentation, Copy-Paste~\cite{dvornik2018modeling, dvornik2019importance, instaboost, cut_paste, wang2019data, lian2021geometry, ghiasi2021simple} is a simple yet effective instance-level data augmentation method. Typically, it copies instance masks of desired category from an image and pastes them to another image. 

PSIS~\cite{wang2019data}, similar to Copy-Paste, generates more training data by switching instances of same class between different images. Dvornik et al.~\cite{dvornik2018modeling, dvornik2019importance} explored global context to guide the placement of instance masks. Similarly, InstaBoost \cite{instaboost} also learns where to paste instances by a probability heatmap. Instead of modeling contextual information to place masks, Cut-Paste \cite{cut_paste} claims that the local adaptation of instance mask itself matters more than the global consistency. More recently, Ghiasi et al.~\cite{ghiasi2021simple} demonstrate that the simple mechanism of pasting objects randomly without any local and global adaptations except for scale jittering is good enough for instance segmentation task. 

In autonomous driving, Copy-Paste is extended for monocular 3D vehicle detection~\cite{lian2021geometry}. The data augmentation approach manipulates camera system to generate new training images with geometric shifts and preserves geometry relationships in objects. We also apply Copy-Paste data augmentation for rare object detection, but with several differences: (1) Data augmentation for rare object detection in autonomous driving has not been explored. In this work, we focus on Copy-Paste data augmentation for rare object (e.g., traffic cones) detection. (2) Existing works commonly ensure the local consistency with gaussion blur and Poisson blending. We show that there is still a huge room for improvement by instance-level local adaptation (e.g., HSV color transform). (3) We propose traffic context aware scaling of local objects and placement of instance masks, in contrast to randomly scaling and placement of objects without any global constraints.
(4) We also present a thorough study of the effectiveness of our local-adaptive and global constraints based Copy-Paste data augmentation for rare object detection.

\section{METHOD}

Fig.~\ref{fig:overview} shows the overview of our Copy-Paste data augmentation method. It consists of 3 major steps: (1) Collection of object instance masks and background images; (2) Determination of where to paste instance masks by understanding traffic scene context and (3) Local-adaptive transform of instance masks. We use traffic cone data augmentation for the explanation in this section.

\subsection{Collecting instance masks and background images}

We collect background images from real traffic scenes. These images covers diverse traffic scenarios such as different road grades, weather conditions, traffic conditions, etc. Examples of background images are shown in Fig.~\ref{fig:dataset}. 

Instance masks are obtained from both online sourcing and real traffic data collection. For example, traffic cones with various types, colors and sizes are collected. In the experiments, we validate that our data augmentation method is not sensitive to different domains of instance masks.

\subsection{Traffic context understanding}

Randomly pasting objects on background images in a context-free manner is not efficient and can result in violation of object relationships in traffic scenes. Taking traffic cone into consideration, in a real traffic scene, traffic cones are usually present on the ground and along lanes in most cases. 
Therefore, random positioning cone masks may lead to generating unreal training images. In our method, we take freespace, traffic lanes and common road users (car, truck, bus, pedestrian, bicycle) to represent the traffic context. To this end, we use a multi-task network to automatically extract these components. Specifically, the multi-task network contains three heads: (1) one semantic segmentation head for freespace segmentation; (2) another instance segmentation head for traffic lane segmentation and (3) a detection head for common road users detection.

In addition, we also take camera parameter as our traffic context. The camera intrinsic parameter is defined by focal length and camera center $(f_x, f_y, c_x, c_y)$. Camera height $h_{\text{cam}}$ is used as camera extrinsic parameter. Camera parameter can be applied to model 3D geometry constraint for the object mask positioning. For example, given a desired distance $d$ to place a traffic cone in the real world, the vertical coordinate $y$ on image space of the superimposition can be simply computed by 

\begin{equation}
     y = f_y \frac{h_{cam}} {d} + c_y
\label{eq_y}
\end{equation}

Our method focuses on the superimposition of instance masks with traffic context as constraints. We summarize the positioning strategies as follows: The traffic cone mask must be placed on freespace region, the traffic cone is likely along traffic lanes and the traffic cone can not be fully occluded with other common road users.

\begin{figure}[t]
	\vspace{2mm}
	\begin{center}
		\begin{tabular}
			{@{\hspace{.0mm}}c@{\hspace{.4mm}} @{\hspace{.4mm}}c@{\hspace{.4mm}}}
			\vspace{-2pt}
			\includegraphics[width=0.45\linewidth]{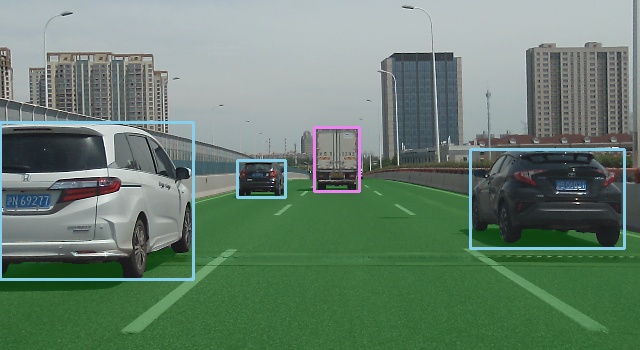}&
			\includegraphics[width=0.45\linewidth]{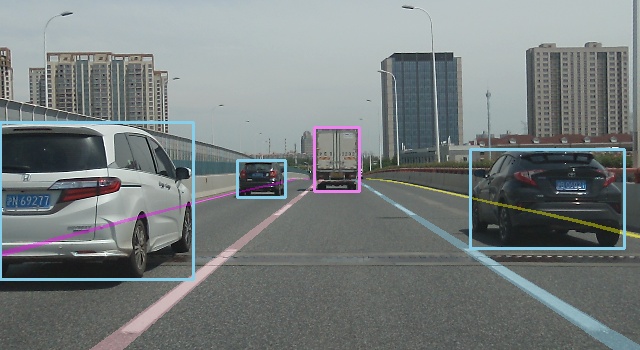} \\
			\vspace{-2pt}
			\includegraphics[width=0.45\linewidth]{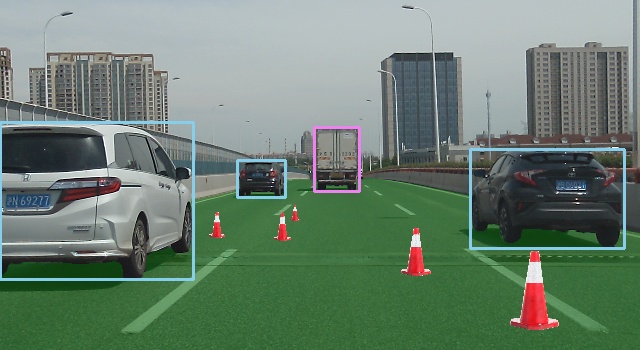}&
            \includegraphics[width=0.45\linewidth]{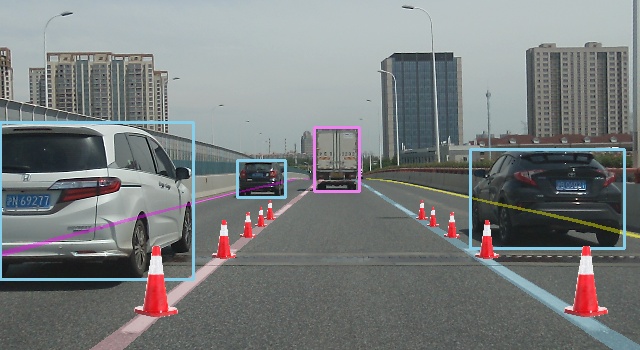} \\
		\end{tabular}
	\end{center}
	{\caption{Traffic scene context aware superimposition of instance masks. First row: Freespace and traffic lanes with common objects computed by a multi-task deep model to obtain the traffic scene understanding. Second row: Depth-aware scaling of instance masks with traffic context constraints.} 
	\label{fig:fig_global_guidance}}
	\vspace{-4mm}
\end{figure}

Fig. \ref{fig:fig_global_guidance} gives some examples of traffic scene context and the final synthetic images.
We demonstrate that, with traffic scene context, these positioning strategies improve the final  performance of the detector.

\subsection{Local-adaptive instance transform}

Instance-level transform of object mask is essential to generate training image that is invariant to local discrepancies. We use the following local-adaptive data augmentation strategies to ensure seamlessly pasting object masks into background images.

\begin{table*}[!t]
    \vspace{2mm}
	\setlength{\tabcolsep}{0.3em}
	\centering
	\footnotesize{
		\caption{We evaluate the effect of components in generating synthetic training images with different settings. $*+$ denotes incremental experimental settings. Common objects are default used in traffic context constraints.}
		\label{tab:main:result} 
		\begin{tabular}{ll|cccc|ccc}
			\hline
			& \multirow{2}{7em}{Methods}   & \multicolumn{4}{c|}{Common objects}  & \multicolumn{3}{c}{Traffic cone}  \\
			& & $\text{mAP}_{\text{50}}$ & $\text{mAP}$ & $\text{Precision}$ & $\text{Recall}$& $\text{AP}$ & $\text{Precision}$ & $\text{Recall}$ \\
			\hline
			Baseline-common & w/o cone & 78.2 & 58.1 & 75.1 & 76.9 & - & - & -\\
			\hline
			Baseline-main & Random~\cite{cut_paste} & 76.3 & 55.7 & 75.8 & 75.0 & 14.7 (+0.0) & 89.0 & 21.8 \\
			\hline
			Traffic Context Aware & Freespace & 76.0 & 55.1 & 75.3 & 74.7 & 13.8 (\hlr{-0.9}) & \textit{\textbf{91.3}} & 21.4 \\
			\hline
			\multirow{4}{7em}{Local-adaptive Instance Transform} & Freespace + Depth-aware scaling &  76.2 & 55.7 & 74.7 & 75.4 & 17.9 (\hlb{+3.2}) & 90.6 & 27.1  \\        
			& $*+$ Color transform (HSV) & 76.0 & 55.5 & 74.7 & 75.5 & 19.4 (\hlb{+4.7}) & 85.4 & 30.2  \\
			& $*+$ Poisson blending & 76.0 & 55.8 & 74.2 & 75.7 & \textit{\textbf{20.2}} (\hlb{+5.5}) & 86.1 & 30.9 \\
			& $*+$ Gaussian blurring & 76.2 & 55.8 & 74.3 & 75.6 & 18.6 (\hlb{+3.9}) & 91.7 & 29.7 \\
			\hline
			\hline
			& Ours & 75.8 & 55.5 & 74.8 & 74.6 & \textbf{21.4} (\hlb{+6.7}) & 84.4 & 33.9 \\
			\hline
		\end{tabular}
	}
\vspace{-4mm}
\end{table*}

\subsubsection{Depth-aware scaling} To ensure a diverse scale coverage, previous methods \cite{dvornik2018modeling, instaboost, ghiasi2021simple} commonly adopt random scaling of object masks. However, this random scaling hurts the 3D geometry relationships between objects. In order to mitigate this shortcoming, we scale pasted objects under the 3D geometric constraint. Precisely, the apparent size of the pasted object in a image depends on its depth to the camera and its size in the real 3D space.

The height $h_{\text{pixel}}$ of an object with real height $h_{\text{real}}$ at a certain depth (distance) $d$ can be inferred by: 
 \begin{equation}
     h_{\text{pixel}} = f_y  \frac{h_{\text{real}}} {d}
 \label{eq_height}
 \end{equation}

We can then scale an instance mask accordingly and paste it with traffic scene context constraints. Fig.~\ref{fig:fig_global_guidance} shows some results of Copy-Paste data augmentation with depth-aware scaling.

\begin{figure}[t]
	\vspace{2mm}
	\begin{center}
		\begin{tabular}
			{@{\hspace{.0mm}}c@{\hspace{.4mm}} @{\hspace{.4mm}}c@{\hspace{.4mm}}}
			\vspace{-2pt}
			\includegraphics[width=0.45\linewidth]{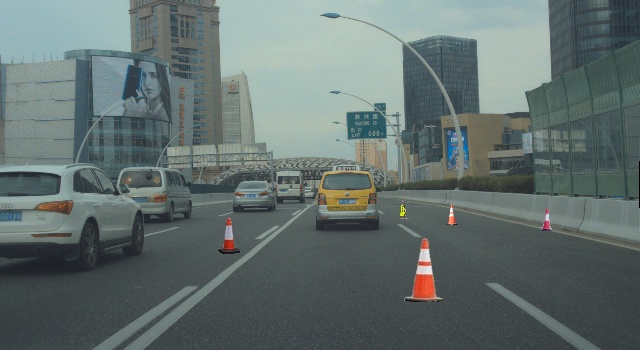}&
			\includegraphics[width=0.45\linewidth]{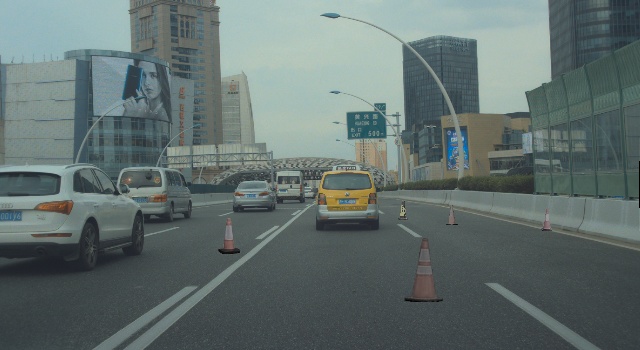} \\
			\vspace{-2pt}
			\includegraphics[width=0.45\linewidth]{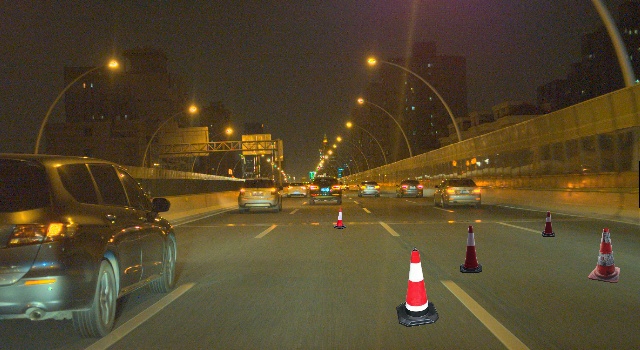}&
			\includegraphics[width=0.45\linewidth]{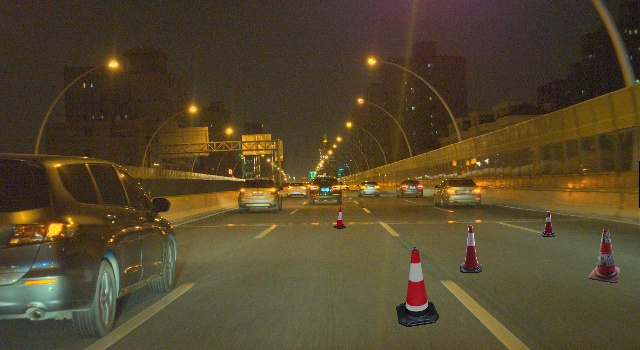} \\
			(a) w/o color transform & (b) w/ color transform 
		\end{tabular}
	\end{center}
	{\caption{Illustration of local-adaptive color transform in HSV space. Local region adaptive color transform yields more realistic images in target domain.} 
	\label{fig_hsv}}
	\vspace{-4 mm}
\end{figure}

\subsubsection{Instance-level blending} Directly pasting objects on background images creates boundary artifacts. To smooth out the seams between pasted objects and background images, we follow \cite{cut_paste, dvornik2018modeling} and apply various blending strategies such as Gaussian blurring and Poisson blending to restrain the artifacts problem. Gaussian smoothing blurs an image by a Gaussian function, mutes image noise and reduces chromatic aberration at high-contrast edges in an image. And Poisson blending removes starkly visible seams and adds lighting variations. We choose to use Poisson blending according to the ablation study. 

\subsubsection{Local-adaptive color transform (HSV)} Previous works mainly focus on the edge smoothness between pasted objects and background images without considering illumination and intensity contrast. 
In real traffic scene, traffic cones may appear at different time (e.g., daytime, evening and night), on various locations (e.g., in shadow of trees, in the dark tunnel or under glare), and during diverse weather conditions (e.g., sunny, cloudy and rainy). Therefore, in these diverse scenarios, the image content varies greatly in brightness, saturation and contrast. 

In order to mitigate the effects of changes in illumination and contrast, we introduce color transform of instance masks to blend them with the background images. Specifically, we use HSV space to serve the color transform. We define $R$ as the local region on the background image to paste the instance mask $M$. Using saturation channel as example, the instance mask saturation is changed by a scale $s = \frac{\hat{M_s}}{\hat{R_s}}$. Here $\hat{R_s}$ and $\hat{M_s}$ denote the mean saturation of region $R$ and instance mask $M$, respectively. Fig.~\ref{fig_hsv} demonstrates that the local region adaptive color transform is critical to generate realistic augmented images.

\section{EXPERIMENTS}

\subsection{New Rare Object Dataset (ROD)}
\begin{figure*}[t]
    \vspace{4mm}
	\begin{center}
		\begin{tabular}
			{@{\hspace{.0mm}}c@{\hspace{.1mm}} 
			 @{\hspace{.1mm}}c@{\hspace{.1mm}}
			 @{\hspace{.1mm}}c@{\hspace{.1mm}}
			 @{\hspace{.1mm}}c@{\hspace{.1mm}}
			 @{\hspace{.1mm}}c@{\hspace{.0mm}}}

			 Night & Rainy & Glare & Large Curve & Traffic Jam \\
			 \vspace{-3pt}
			\includegraphics[width=0.19\linewidth]{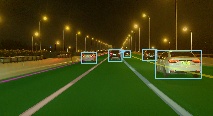}&
			\includegraphics[width=0.19\linewidth]{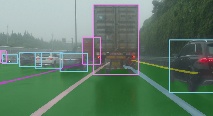}&
			\includegraphics[width=0.19\linewidth]{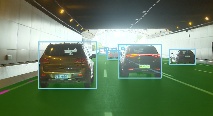}&
			\includegraphics[width=0.19\linewidth]{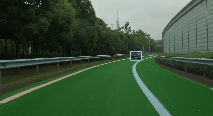}&
			\includegraphics[width=0.19\linewidth]{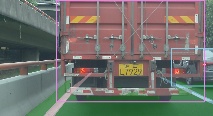} \\
			\vspace{-3pt} 
			\includegraphics[width=0.19\linewidth]{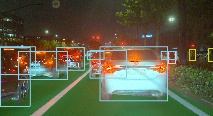}&
			\includegraphics[width=0.19\linewidth]{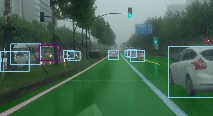}&
			\includegraphics[width=0.19\linewidth]{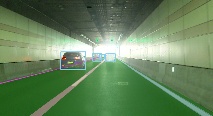}&
			\includegraphics[width=0.19\linewidth]{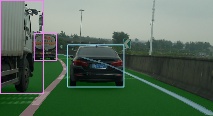}&
			\includegraphics[width=0.19\linewidth]{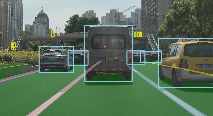} \\
			\vspace{-3pt}
			\includegraphics[width=0.19\linewidth]{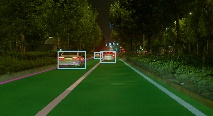}&
			\includegraphics[width=0.19\linewidth]{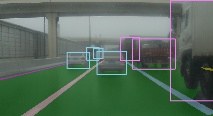}&
			\includegraphics[width=0.19\linewidth]{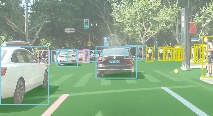}&
			\includegraphics[width=0.19\linewidth]{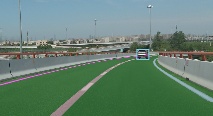}&
			\includegraphics[width=0.19\linewidth]{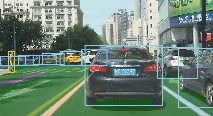} \\
		\end{tabular}
	\end{center}
	
	\vspace{-2 mm}

	{\caption{Overview of our newly released Rare Object Dataset (ROD). It covers a diverse set of scenarios under various road grades, weather conditions, time, and scenes.}
	\label{fig:dataset}}
\vspace{-2 mm}
\end{figure*}

We have built and released a new Rare Object Dataset (\textbf{ROD})~\footnote{\url{https://nullmax-vision.github.io}}. The dataset contains 10k training images, 4k validation images and the corresponding 2D bounding box annotations with 7 representative object categories (car, truck, bus, pedestrian, bicycle, traffic cone and traffic barrel). The dataset also covers a diverse range of scenarios, such as different road grades (e.g. highway, expressway, city street and country road), different weathers (e.g., sunny, cloudy and rainy) and different times of the day (e.g. daytime, evening and night). In addition, we also make 1k traffic cone masks, 100 traffic barrel masks, and 50 traffic warning triangle masks available to the community. Fig.~\ref{fig:dataset} gives some typical examples of the ROD. Also, experimental results in Table.~\ref{tab:main:result} shows that our dataset for rare object detection (traffic cone) is very challenging due to the diversity of the dataset.

\subsection{Experimental Settings}
\textbf{Architecture and Training Details.} We use Yolov5\cite{glenn_jocher_2020_4154370} as our baseline detector.
All models are trained using a batch size of 30 per GPU and weight decay of 5e-4. We use a learning rate of 1e-2 and gradually decay it to 1e-3. EMA strategy is taken to smooth the network parameters. All the models are trained from scratch with 200 epochs. For fair comparison, in each experimental setting, a background image is pasted with 5 transformed mask in the training.

\begin{figure}[t]
	\begin{center}
		\begin{tabular}
			{@{\hspace{.0mm}}c@{\hspace{.4mm}} @{\hspace{.4mm}}c@{\hspace{.4mm}}}
			\vspace{-2pt}
			\includegraphics[width=0.45\linewidth]{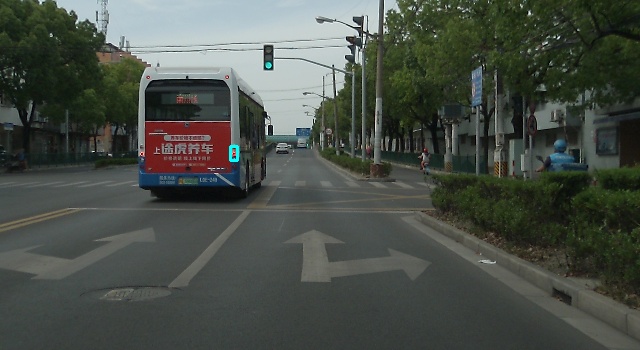}&
			\includegraphics[width=0.45\linewidth]{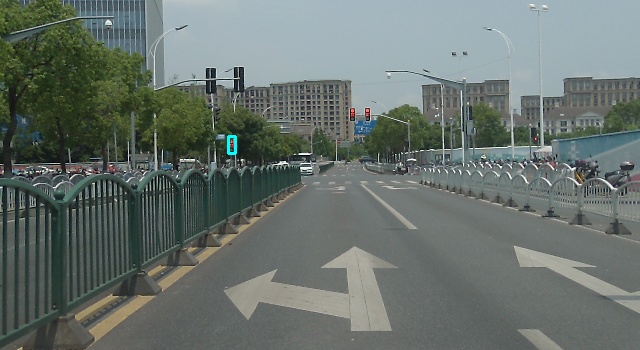} \\
			\vspace{-2pt}
			\includegraphics[width=0.45\linewidth]{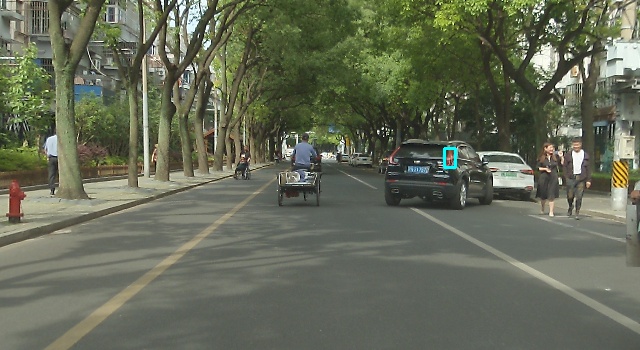}&
            \includegraphics[width=0.45\linewidth]{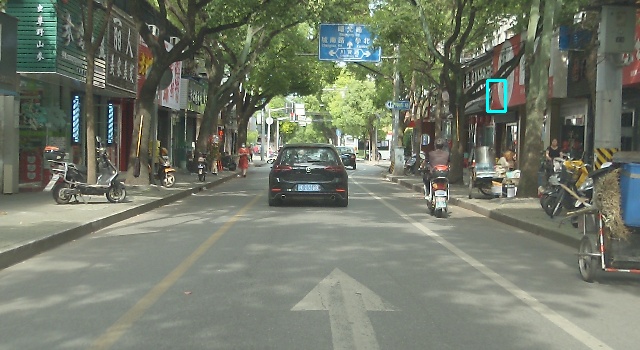} \\
		\end{tabular}
	\end{center}
	{\caption{Traffic cone detection results without freespace constraint. False positives are shown in rectangles.} 
	\label{fig:fig_failure}}
\end{figure}

\subsection{Evaluation metrics and results}
To investigate the performance of our approach, we follow the convention in object detection task and use $\text{mAP}$, $\text{Precision}$ and $\text{Recall}$ as the evaluation metrics.

We train a baseline for common road users (car, truck, bus, pedestrian, bicycle) and denote as Baseline-common. Following~\cite{cut_paste}, we train a detector by randomly pasting instance masks into background images. This detector is used as the main baseline. We study the effect of components in generating synthetic training images with different settings.
Experimental results of training with synthetic images and evaluated on real annotated data are shown in Table.~\ref{tab:main:result}. 

With Copy-Paste data augmentation, we achieve slight performance drop of common object detection and get desirable results on traffic cone detection. Compared to standard Copy-Paste~\cite{cut_paste} which randomly pasting the masks on background images, our traffic context (freespace) aware data augmentation yields better result of traffic cone detection on $\text{Precision}$. Some detection results are shown in Fig.~\ref{fig:fig_failure}. We can observe that randomly pasting instances without traffic context constraints gives lots of false positives. For example, cases including taillights, traffic lights or artifacts on sky are usually recognized as traffic cones. However, using freespace as constraint, the traffic cone can only be placed on the ground. Therefore, our data augmentation method helps reduce false positives on regions that traffic cones unlikely to be present.

With local-adaptive instance transform, the traffic cone detection performance is further improved. Table.~\ref{tab:main:result} validates the effect of each component of local-adaptive transform component. In addition, we also show that Poisson blending gives better results than Gaussion blending.

In summary, our data augmentation method which combines local-adaptive instance transform and traffic context constraints boosts the performance of traffic cone detection.

\subsection{Ablation Study}

In the ablation study, we create a subset with 5K images to conduct followings experiments.

\textbf{Does domain of source masks matters?} Generally, directly transferring instance masks from source domain and applying them to another different target domain may degrade the detection performance. Thanks to our introduced local-adaptive instance transform, our data augmentation method is not sensitive to domain of instance masks. Table.~\ref{tab:ablation_domain} shows that training with data generated from different domains achieves comparable results.

\begin{table}[t]
\small
\centering
\caption{Sensitivity of source domain of instance masks. We show that our data augmentation method is not sensitive to source domain of instance masks.}
\begin{tabular}{cccccc}
	\hline
	Domain & \# masks & $\text{AP}$ & $\text{Precision}$ & $\text{Recall}$ \\
  \hline
  Online & 90 & 18.6 & 69.7 & 33.1 \\
  ROD & 90 & 18.3 & 72.9 & 32.8  \\
  \hline
\end{tabular}
\label{tab:ablation_domain}  
\end{table}

\textbf{How many masks do we need to collect?}
We conduct an experiment to validate the effect of number of instance masks images to generating training images. From Table.~\ref{tab:ablation_numbers}, we conclude that one can use few representative instance masks to obtain competitive detection performance. We think that our data augmentation covers a variety of scenarios and contributes to the findings. For example, although we do not have a cone mask at night, our local adaptive color transform is able to generate masks at various illumination conditions.

\begin{table}[t]
\small
\centering
\caption{Evaluation of number of instance masks to generate training images and number of augmented instances (Aug) for training a cone detector. Evaluation on ROD by varying the number of real data and synthetic training images.}
\begin{tabular}{ccccc}
	\hline
	\hline 
	&  Common objects & \multicolumn{3}{c}{Traffic cone}  \\
	& $\text{mAP}_{\text{50}}$ & $\text{AP}$ & $\text{Precision}$ & $\text{Recall}$ \\
	\hline
	\hline
	\# Masks &  \\ 
  \hline
  100 &  72.6 & \textbf{22.5} & 71.3 & 40.3 \\
  300 &  72.5 & 20.0 & 76.4 & 36.5 \\
  500 &  71.9 & 21.2 & 72.4 & 39.4 \\
  700 &  72.8 & 20.9 & 69.7 & 37.4 \\
  900 &  72.2 & 19.9 & 74.5 & 36.4 \\
  \hline
  \hline
  \# Aug & \\ 
  \hline
  5000 & 71.4 & 12.8 & 78.2 & 22.2 \\
  15,000 & 69.0 & 16.0 & 74.5 & 28.1 \\
  25,000 & 70.0 & 18.6 & 69.7 & 33.1 \\
  35,000 & 69.1 & \textbf{20.9} & 72.2 & 38.1 \\
  45,000 & 68.3 & 19.2 & 69.6 & 35.6 \\
  \hline
  \hline
	Dataset & \\ 
  \hline
  Real & 72.8 & \textit{\textbf{19.2}} & 78.0 & 38.8 \\
  Syn & 70.0 & 18.6 & 69.7 & 33.1 \\
  Real + Syn & 73.0 & \textbf{28.12} & 71.13 & 51.83\\
  \hline
\end{tabular}
\label{tab:ablation_numbers}  
\end{table}

\textbf{How many instances do we need to augment?}
Similarly, Table.~\ref{tab:ablation_numbers} shows results of the effect of number of augmented instances to train a detector. We use 90 traffic cones for this experiment. Our data augmentation method is efficient to generate effective training images. We draw that one can use considerably few augmented instances on background images to obtain comparable detection performance. 

\begin{table}[htbp]
\small
\centering
\caption{Generalization of our data augmentation method on traffic barrel detection.}
\begin{tabular}{cccc}
	\hline
	Dataset & $\text{AP}$ & $\text{Precision}$ & $\text{Recall}$ \\
 \hline
  Syn  & 28.5 & 86.7 & 60.0 \\
  \hline
\end{tabular}
\label{tab:ablation_extend_barrel}  
\end{table}

\textbf{Evaluation of Real vs Synthetic dataset}
Table.~\ref{tab:ablation_numbers} presents results of using real and synthetic training data with different settings. It can be observed that by using only 90 instance masks to generate synthetic training images, we can get competitive performance compared to using real annotated training data. Also, combing both real and synthetic images, the detection performance is significantly boosted.

\textbf{Generalization to other rare object detection tasks.}
To validate generalization of our data augmentation method on other type rare object detection, we present experiments on traffic barrel detection task only using synthetic training images. Table.~\ref{tab:ablation_extend_barrel} shows that our proposed data augmentation approach can be easily applied on other rare object detection task in autonomous driving.

\section{CONCLUSIONS}

In this paper, we propose an empirical study of Copy-Paste data augmentation for rare object detection in autonomous driving. Our method consists of local-adaptive instance level transform of object masks and superimposition of masks under guidance of global traffic context information. We show that simple data augmentation method performs well on rare object detection. In addition, we also demonstrate that local-adaptive transform of object masks is essential for mask multiplications. In future work, we would like to explore rare object detection in incremental training setting.

\vspace{0mm}













\bibliographystyle{IEEEtran}
\bibliography{refs}


\end{document}